\documentclass[lettersize,journal]{IEEEtran}
\usepackage{amsmath,amsfonts}
\usepackage{cite}

\usepackage{algorithmic}
\usepackage{algorithm}
\usepackage{url}
\usepackage{orcidlink}
\usepackage{mystyle}
\newcommand{\red}[1]{{\color{red}#1}}
\begin{document}

\title{Hyperspectral Remote Sensing Images \\ Salient Object Detection: \\ The First Benchmark Dataset and Baseline}

\author{
Peifu Liu$^*$, Huiyan Bai$^*$, Tingfa Xu$^{\dagger}$, Jihui Wang$^{\dagger}$, Huan Chen, Jianan Li$^{\dagger}$
    \thanks{Peifu Liu, Huiyan Bai, Tingfa Xu, Jihui Wang, Huan Chen, and Jianan Li are with Beijing Institute of Technology, Beijing 100081, China, and with the Key Laboratory of Photoelectronic Imaging Technology and System, Ministry of Education of China, Beijing 100081, China. Email: \{3120245389, 3120220533, ciom\_xtf1, wjhzhaojie, 3220235096, lijianan\}@bit.edu.cn.} %
    \thanks{Tingfa Xu is also with the Big Data and Artificial Intelligence Laboratory, Beijing Institute of Technology Chongqing Innovation Center, Chongqing 401151, China.
    } %
    \thanks{$^*$ Equal Contribution.}
    \thanks{$^{\dagger}$ Correspondence to: Tingfa Xu, Jihui Wang and Jianan Li.}
}

\markboth{Journal of \LaTeX\ Class Files,~Vol.~14, No.~8, August~2021}%
{Shell \MakeLowercase{\textit{et al.}}: A Sample Article Using IEEEtran.cls for IEEE Journals}


\maketitle

\begin{abstract}
The objective of hyperspectral remote sensing image salient object detection (HRSI-SOD) is to identify objects or regions that exhibit distinct spectrum contrasts with the background. This area holds significant promise for practical applications; however, progress has been limited by a notable scarcity of dedicated datasets and methodologies. To bridge this gap and stimulate further research, we introduce the first HRSI-SOD dataset, termed HRSSD, which includes 704 hyperspectral images and 5327 pixel-level annotated salient objects. The HRSSD dataset poses substantial challenges for salient object detection algorithms due to large scale variation, diverse foreground-background relations, and multi-salient objects.
Additionally, we propose an innovative and efficient baseline model for HRSI-SOD, termed the Deep Spectral Saliency Network (DSSN). The core of DSSN is the Cross-level Saliency Assessment Block, which performs pixel-wise attention and evaluates the contributions of multi-scale similarity maps at each spatial location, effectively reducing erroneous responses in cluttered regions and emphasizes salient regions across scales. 
Additionally, the High-resolution Fusion Module combines bottom-up fusion strategy and learned spatial upsampling to leverage the strengths of multi-scale saliency maps, ensuring accurate localization of small objects.
Experiments on the HRSSD dataset robustly validate the superiority of DSSN, underscoring the critical need for specialized datasets and methodologies in this domain. Further evaluations on the HSOD-BIT and HS-SOD datasets demonstrate the generalizability of the proposed method. The dataset and source code are publicly available at https://github.com/laprf/HRSSD.
\end{abstract}

\begin{IEEEkeywords}
Salient Object Detection, Hyperspectral Remote Sensing Images, Spectral Saliency, HRSSD Dataset
\end{IEEEkeywords}

\section{Introduction}
\begin{figure}[tp]
  \centering
  \includegraphics[width=\linewidth]{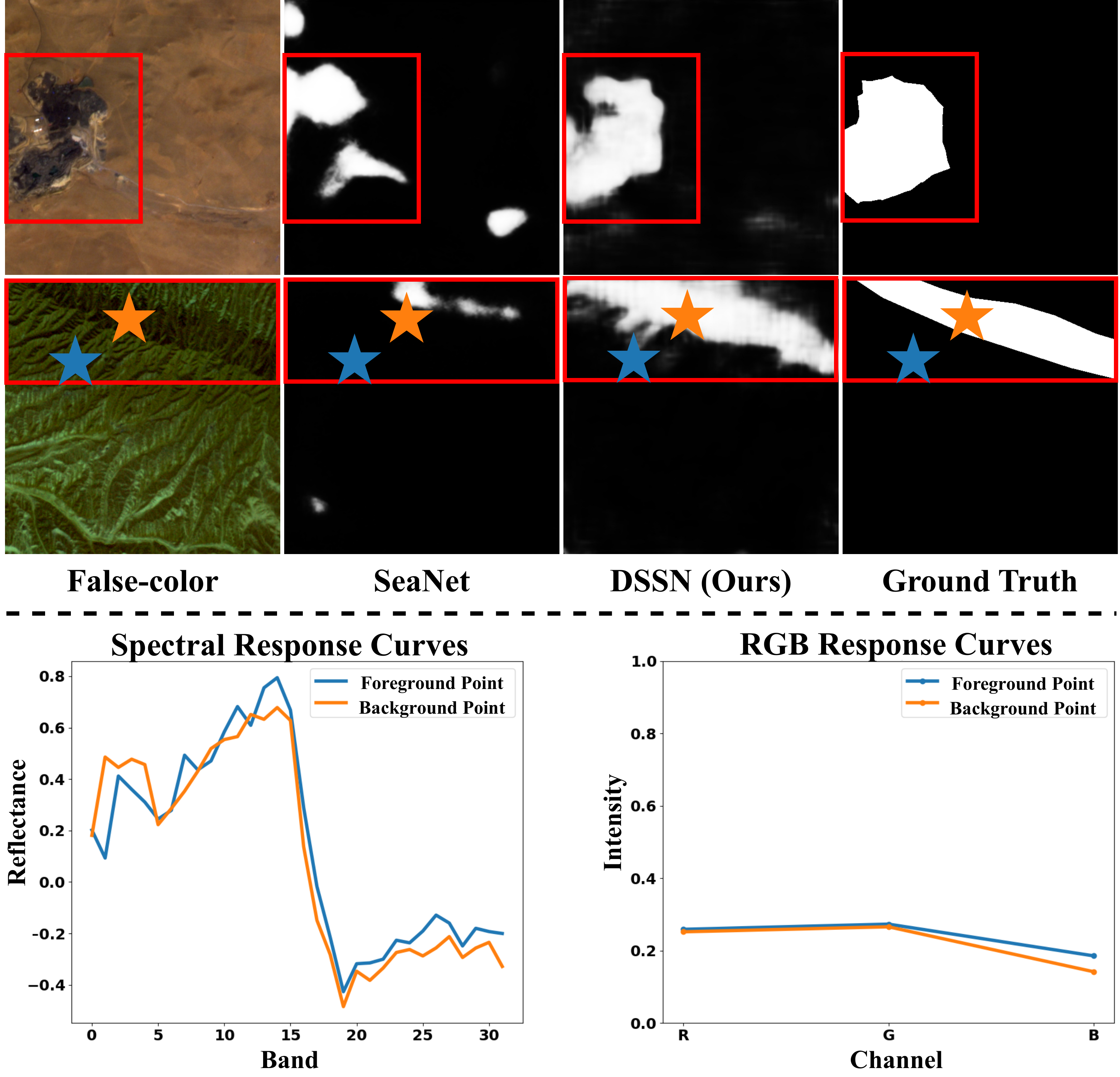}
  \caption{Comparison with SeaNet, an RGB image-based method that struggles to detect salient objects when foreground (orange star) and background (blue star) colors are similar due to nearly uniform channel responses. In contrast, DSSN leverages rich spectral information for improved detection. The false-color image is synthesized by mapping three spectral bands (730 nm, 580 nm, and 466 nm) to the red, green, and blue channels, respectively.
}
  \label{Fig: motivation}
\end{figure}

\IEEEPARstart{H}{yperspectral} salient object detection (HSOD) aims to identify objects or regions within a scene that exhibit distinct spectral characteristics compared to their surroundings~\cite{SMN, qin2024dmssn, SPSD}. Hyperspectral images (HSIs) capture detailed spectral reflectance properties, enabling differentiation of materials based on their unique spectral signatures~\cite{10261266,zhao_TCI}. This capability provides HSIs with a significant advantage over RGB images, particularly in scenarios where foreground (orange star) and background (blue star) objects share similar visual appearances but differ spectrally~\cref{Fig: motivation}.

Remote sensing images, acquired from satellites, drones, or other spacecraft, offer a bird's-eye view of Earth's surface. Despite the various application scenarios for hyperspectral remote sensing image salient object detection (HRSI-SOD), such as military defense~\cite{Khan2018Modern}, mineralogical mapping~\cite{Sandra2021Feature}, and atmospheric monitoring~\cite{liu2022first}, research in this domain has predominantly focused on natural scenes~\cite{Imamoglu2018HSSOD, SMN, qin2024dmssn, imamouglu2019salient, SPSD}. This trend can be attributed to two main factors: \textbf{(i)}  the scarcity of publicly available datasets tailored for HRSI-SOD, and \textbf{(ii)}  the lack of specialized methodologies addressing the unique challenges of this domain.

To advance this field, we introduce the Hyperspectral Remote Sensing Saliency Dataset (HRSSD) , the first benchmark dataset specifically designed for HRSI-SOD. Derived from WHU-OHS~\cite{Li2022WHUOHS}, HRSSD comprises 704 hyperspectral images, each with a spatial resolution of $512 \times 512$ pixels and 32 spectral channels spanning the 466-940~nm wavelength range, accompanied by pixel-level annotations for 5327 salient objects. This dataset's broad coverage and varied land cover types present three primary challenges for salient object detection: \textbf{(i) Large Scale Variation:} Spatial scales of land cover differ substantially, leading to significant variation in object sizes across images. Additionally, the spatial distribution of the same land cover type varies widely across regions, increasing the diversity of salient object sizes within a single image. \textbf{(ii) Diverse
Foreground-background Relations:} The spatial distribution and spectral characteristics of land cover, combined with its differing roles across various environments, render certain land covers salient in specific scenes, while they appear as background in others, complicating foreground-background delineation. \textbf{(iii) Multi-salient Objects:} The diversity and uneven spatial distribution of land cover types result in numerous images containing multiple salient objects, increasing detection complexity. These challenges provide a strong basis for developing and evaluating advanced HRSI-SOD methods.

The proposed methodology is motivated by the fundamental principle of "center-surround spectral difference evaluation," which evaluates the spectral similarity between central pixels and their surrounding neighbors to identify salient objects. Traditional methods~\cite{itti_1998_a, Liang2013SODHS} predominantly rely on this principle. For instance, Spectral Saliency~\cite{Liang2013SODHS} constructs a pyramid structure by downsampling the input hyperspectral image and calculates spectral angle distances between pyramid layers to generate similarity maps. The final saliency map is produced by directly summing these individual maps. While this approach effectively implements "center-surround spectral difference evaluation," it suffers from three critical limitations: \textbf{(i) Limited Multi-Scale Adaptability:} Deep downsampling results in the loss of small targets, and the direct summation of multi-scale similarity maps often causes larger targets to dominate, obscuring smaller ones. \textbf{(ii) Inadequate Foreground-Background Discrimination:} Relying solely on spectral information is insufficient for distinguishing foreground objects from background regions when their spectral characteristics are similar. \textbf{(iii) Suboptimal Handling of Multiple Objects:} Direct summation of similarity maps causes interference among multiple salient objects, leading to merged detections or missed small, scattered targets due to a lack of robust multi-object separation mechanisms.

To overcome these limitations, we propose the Deep Spectral Saliency Network (DSSN), an innovative baseline model for HRSI-SOD, which consists of three parts: Spatial-spectral Joint Feature Extractor (SJFE), Cross-level Saliency Assessment Block (CSAB), and High-resolution Fusion Module (HRFM). Each component is specifically designed to tackle one or more core limitations of traditional methods, enabling DSSN to robustly address large-scale variation, diverse foreground-background relations, and multi-salient object challenges in HRSI-SOD.

The Spatial-spectral Joint Feature Extractor enhances foreground-background discriminability through parallel spatial and spectral branches. The spatial branch captures fine-grained textural details, while the spectral branch employs a Spectral Attention Module to fully exploit spectral information. A hierarchical fusion mechanism adaptively integrates spatial and spectral features via cross-branch modulation, strengthening the representation of spatial-spectral features and improving foreground-background differentiation.

As the core structure of DSSN, the Cross-level Saliency Assessment Block effectively implements "center-surround spectral difference evaluation". It first generates initial similarity maps by measuring cross-layer feature correlations. Then, it applies pixel-wise attention mechanisms to evaluate the contributions of multi-scale similarity maps at each spatial location, enabling adaptive fusion through weighted aggregation. This fusion strategy suppresses erroneous responses in cluttered regions and emphasizes salient regions across scales, inherently addressing challenges related to large-scale variations and multiple salient objects.

The High-resolution Fusion Module integrates multi-scale intermediate saliency maps using a bottom-up fusion strategy, progressively combining high-resolution saliency maps with lower-resolution ones. This strategy leverages the strengths of each scale: high-resolution maps capture small, scattered objects, while low-resolution maps capture larger ones. Additionally, learned spatial upsampling is employed to recover fine-grained details, ensuring accurate localization of small objects. This design ensures that DSSN can robustly detect multiple salient objects of varying sizes, even when they are closely spaced.

Experiments conducted on the HRSSD dataset underscore the critical importance of introducing a specialized benchmark for this domain. The results also highlight the high accuracy and computational efficiency of our DSSN. Further evaluations on additional datasets, including HSOD-BIT~\cite{qin2024dmssn} and HS-SOD~\cite{Imamoglu2018HSSOD}, confirm the strong generalizability of our approach. In a nutshell, our contributions can be summarized as follows:

\begin{itemize}
 \item We establish the first benchmark dataset for HRSI-SOD, introducing challenges such as large scale variation, diverse foreground-background relations, and multi-salient objects.

 \item We propose a pioneering baseline model that specifically designed for hyperspectral remote sensing images salient object detection.

 \item We introduce two innovative modules-CSAB and HRFM-that achieves accurate detection performance through pixel-wise attention mechanism and bottom-up multi-scale fusion strategy, respectively.
\end{itemize}

\begin{table*}[tp]
    \centering
    \caption{Comparison between our HRSSD dataset and the main salient object detection datasets. Statistical metrics include the number of images (\#Images), the number of salient objects per image (\#Objects), and image resolution.}
    \label{Tab: Datasets}
    \setlength{\tabcolsep}{6.5mm}
    \begin{tabular}{l|cccccc}
    \toprule[1.2pt]
        Dataset & Data & Scene & Year & \#Images & \#Objects & Resolution \\
        \midrule
        ECSSD~\cite{Shi2016ECSSD} & \multirow{5}{*}{Optical} & \multirow{5}{*}{Natural} & 2013 & 1000 &  $ \sim 1 $ &  $ \sim 400 \times 300 $ \\
        DUT-OMRON~\cite{Yang2013DUT} & & & 2013 & 5168 & $ \sim 5 $ & $ \sim 400 \times 400 $ \\
        PASCAL-S~\cite{Li2014PASCAL} & & & 2014 & 850 & $ \sim 5 $ & Variable \\
        HKU-IS~\cite{Li2015HKU} & & & 2015 & 4447 & Multiple &  $ \sim 400 \times 300 $ \\
        DUTS-TE~\cite{Wang2017DUTS} & & & 2017 & 15572 & Multiple &  $ \sim 400 \times 300 $ \\
        \midrule
        ORSSD~\cite{Li2019ORSSD} & \multirow{5}{*}{Optical} & \multirow{5}{*}{\parbox{1cm}{\centering Remote \\ Sensing}} & 2019 & 800 & Multiple & Variable \\
        EORSSD~\cite{Zhang2021EORSSD} & & & 2020 & 2000 & Multiple & Variable \\
        ORSI-4199~\cite{Tu2022ORSI4199} & & & 2022 & 4199 & Multiple & Variable \\
        RSISOD~\cite{Zheng2023RSISOD} & & & 2023 & 5054 & Multiple & Variable \\
        RSSOD~\cite{Xiong2023RSSOD} & & & 2023 & 6000 & Multiple & Variable \\
        \midrule
         HS-SOD~\cite{Imamoglu2018HSSOD} & \multirow{2}{*}{Hyperspectral} & \multirow{2}{*}{Natural} & 2018 & 60 & 1 & $768 \times 1024$ \\
         HSOD-BIT~\cite{qin2024dmssn} & & & 2024 & 319 & Multiple & $1240 \times 1680$ \\
        \midrule
        \rowcolor{Gray} 
         HRSSD (Ours) & Hyperspectral & \parbox{1cm}{\centering Remote \\ Sensing} & 2025 & 704 & Multiple & $512 \times 512$ \\
    \bottomrule[1.2pt]
    \end{tabular}
\end{table*}

\section{Related Work}
\subsection{Salient Object Detection Datasets}
Significant progress in salient object detection (SOD) has been driven by benchmark datasets with pixel-level annotations. For optical natural scenes, five predominant datasets are widely adopted: ECSSD~\cite{Shi2016ECSSD}, DUT-OMRON~\cite{Yang2013DUT}, PASCAL-S~\cite{Li2014PASCAL}, HKU-IS~\cite{Li2015HKU}, and DUTS-TE~\cite{Wang2017DUTS}. The development of SOD in optical remote sensing imagery remains nascent. Initial efforts include ORSSD~\cite{Li2019ORSSD} (800 images), followed by expanded collections like EORSSD~\cite{Zhang2021EORSSD} (2,000 images) and ORSI-4199~\cite{Tu2022ORSI4199} (4,199 images). Recent datasets RSISOD~\cite{Zheng2023RSISOD} and RSSOD~\cite{Xiong2023RSSOD} specifically address challenges of scale variation, cluttered backgrounds, and seasonal diversity. However, these datasets fail to highlight the advantages of hyperspectral images in detecting salient objects.

In 2018, Imamoglu~\cite{Imamoglu2018HSSOD} introduced the pioneering HS-SOD dataset, tailored for HSOD. The dataset does not include hsi-advantageous scenarios and presents few challenges. In 2024, Qin~\cite{qin2024dmssn} proposed the HSOD-BIT dataset to address these issues. This dataset contains 319 natural scene images with foreground-background color similarity, overexposure, and uneven illumination challenges. Notably, no existing dataset targets salient object detection in hyperspectral remote sensing images. To fill this gap, we present the first benchmark specifically designed for HRSI-SOD, namely HRSSD. Key dataset characteristics are compared in~\cref{Tab: Datasets}.

\subsection{Hyperspectral Salient Object Detection}
Spectral saliency refers to the visual representation formed by aggregating pixels in HSIs with similar spectral characteristics that are visually distinguishable. This concept was first formalized by Le Moan~\etal~\cite{Le_2013_Saliency}. Early methods primarily built upon Itti's visual attention model~\cite{itti_1998_a}. For instance, Liang~\etal~\cite{Liang2013SODHS} proposed a spectral pyramid approach that calculates spectral similarity through Euclidean or spectral angle distance metrics across multiple scales. Despite their computational efficiency, these methods are limited by hand-crafted features that struggle in complex scenes.

The advent of convolutional neural networks (CNNs) with advanced feature extraction capabilities has revolutionized this field. Imamoglu~\etal~\cite{imamouglu2019salient} and Huang~\etal~\cite{huang_2021_salient} integrated CNNs with single/dual-branch architectures and clustering techniques (\eg manifold ranking) to improve accuracy. To address clustering limitations, Liu \etal~\cite{SMN} proposed SMN, the first end-to-end trainable deep network. SMN incorporates specialized modules to extract high- and low-frequency components, along with Mixed-frequency Attention for enhanced feature utilization. In an effort to reduce annotation costs, Liu~\etal \cite{SPSD} proposed the first weakly supervised pipeline for salient object detection in hyperspectral images. However, these methods focus on natural scenes, with no existing solutions specifically addressing remote sensing scenarios.

\subsection{Optical Remote Sensing Images Salient Object Detection}
Concurrent with HSOD advancements, optical remote sensing image salient object detection (ORSI-SOD) has undergone transformative evolution. Early ORSI-SOD methods relied on hand-crafted features that inherently limit detection robustness. The emergence of deep learning has fundamentally reshaped this landscape, with encoder-decoder architectures becoming the dominant paradigm. Landmark contributions include LV-Net~\cite{Li2019ORSSD} - the pioneering end-to-end fully convolutional framework with dual-stream pyramid modules for multi-scale feature fusion. Subsequent innovations have focused on architectural enhancements. For instance, Li~\etal~\cite{Li2023ACCoNet} developed an adjacent context coordination module combining local saliency highlighting and global context integration from adjacent levels. Li~\etal~\cite{Li2022CorrNet} introduced a correlation module with coarse-to-fine strategy, leveraging high-level semantic guidance for low-level feature localization. Li~\etal~\cite{Li2023SeaNet} proposed dynamic semantic matching and edge self-alignment modules for cross-level feature processing. Gao~\etal~\cite{10038723} pioneered Transformer integration through adaptive spatial tokenization encoders.

Despite these advances, existing ORSI-SOD methods remain incompatible with hyperspectral data. Bridging this domain gap, our proposed DSSN establishes the first dedicated baseline for HRSI-SOD. By preserving "center-surround spectral difference" principle, DSSN provides a foundational framework to advance this under-explored field.

\section{HRSSD Benchmark Dataset}
\subsection{Dataset Construction}
The HRSSD dataset was curated by carefully selecting 704 images from the WHU-OHS dataset~\cite{Li2022WHUOHS}, containing 5327 pixel-level annotated salient objects. The dataset is split into training and testing sets in a 5:2 ratio. Below, we provide a concise overview of the dataset construction process.

\noindent\textbf{WHU-OHS Dataset.}
The WHU-OHS dataset is a comprehensive resource for land-cover classification using hyperspectral images, comprising 7795 image patches captured by the Orbita hyperspectral satellite across over 40 locations in China. Covering an area exceeding $150000~km^2$, each patch measures $512 \times 512$ pixels and includes 32 spectral bands spanning 466-940~nm. The dataset is labeled with 25 classes (0-24), where label 0 denotes ignored pixels, and labels 1-24 correspond to distinct land-cover categories. Our dataset retains the core attributes of WHU-OHS, and its authors have approved our reuse and redistribution.

\noindent\textbf{Data Processing.}
In the original dataset, some images are segmented into training and testing subsets. We reassembled these segments to restore image integrity. Additionally, due to potential angular deviations in satellite-captured imagery, certain regions exhibit atypical radiance values. To mitigate this, we generated pixel-level masks for skewed regions, setting their pixel values to zero. This approach minimizes the influence of outliers, enhancing detection accuracy.

\noindent\textbf{Data Filtering.}
We designate the category with the highest pixel count as the background, with all other categories classified as the foreground. Images were filtered based on spatial and spectral characteristics, followed by expert annotator voting.

\noindent\textit{Step 1: Spatial Contrast Filtering.}
Spatial contrast refers to the size difference between foreground and background. We prioritized images with significant spatial contrast and clear foreground-background boundary delineation.

\begin{figure}[tp]
    \centering
    \includegraphics[width=\linewidth]{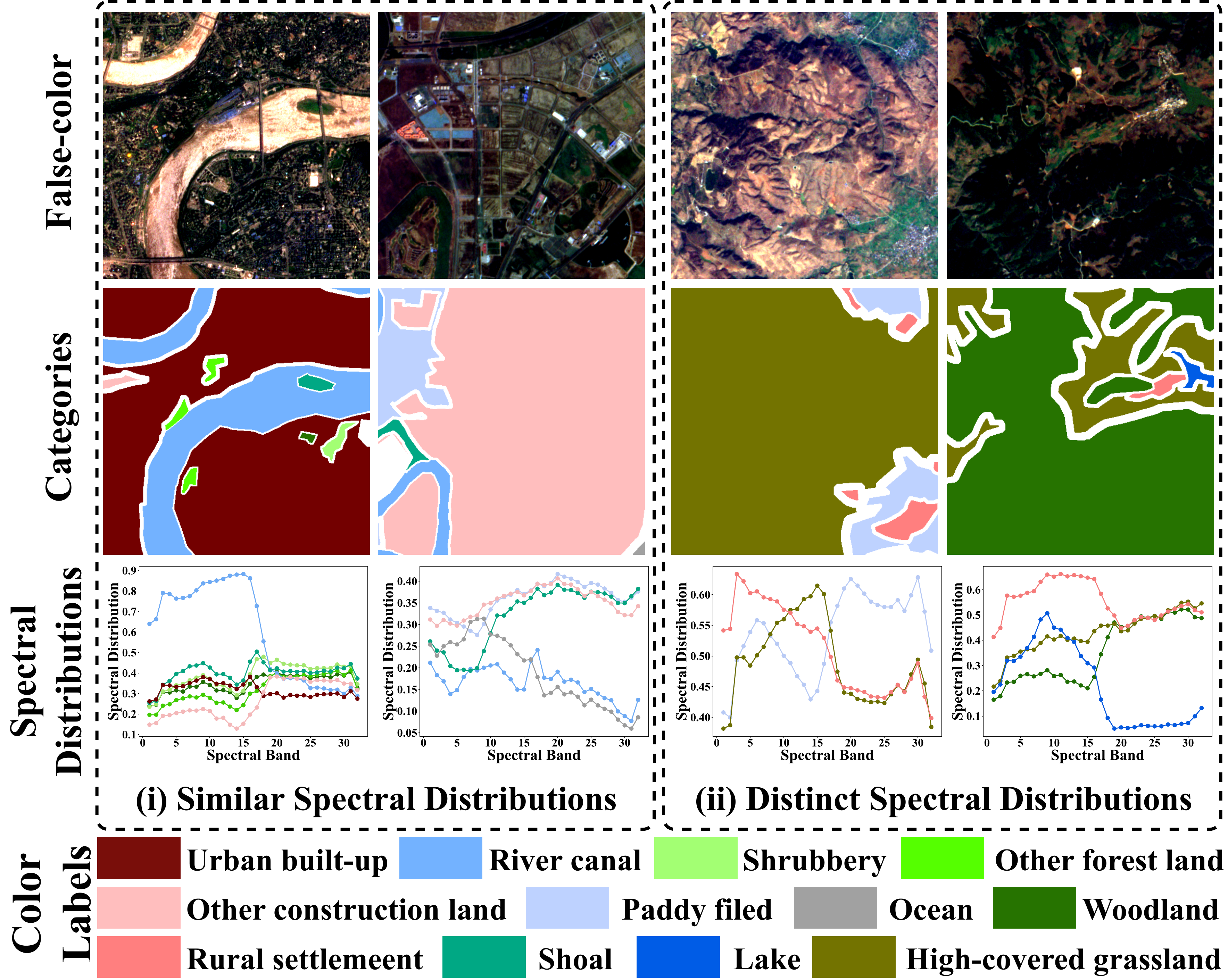}
    \caption{Visualization of spectral statistics. To ensure significant spectral differences between the foreground and background, we excluded images with \textbf{(i)} similar spectral distributions and retained those with \textbf{(ii)} distinct spectral distributions across different land-cover categories.}
    \label{Fig: Spectral distribution}
\end{figure}

\noindent\textit{Step 2: Spectral Differentiation Filtering.}
Effective hyperspectral salient object detection requires substantial spectral contrast between foreground and background. We conducted spectral analysis to exclude images where spectral responses were too similar, as shown in~\cref{Fig: Spectral distribution}~\red{(i)}, and retained those with distinct spectral differences, as in~\cref{Fig: Spectral distribution}~\red{(ii)}.

\noindent\textit{Step 3: Voting-based Filtering.}
Annotators underwent three hours of training before participating in the filtering task. An image was included only if at least three out of five annotators agreed it met the criteria. Through this process, 806 images were initially selected, and after further expert review, 704 images were finalized to form the HRSSD dataset. The resulting dataset features distinct spectral differences between foreground and background, with foreground objects meeting salient object size requirements.

\noindent\textbf{Salient Object Annotation.}
Background pixels are labeled as 0, foreground pixels as 1, and "ignore" pixels (e.g., boundaries with spectral mixing) as -1. Using this labeling strategy, we generated ground-truth images for the dataset.

\subsection{Statistics and Analysis}
Based on the HRSSD dataset, we provide a comprehensive statistical analysis of salient object counts, object sizes, size bias, and center bias. These metrics enhance understanding of the dataset's characteristics and its challenges.

\begin{figure*}[tp]
  \centering
  \includegraphics[width=0.9\linewidth]{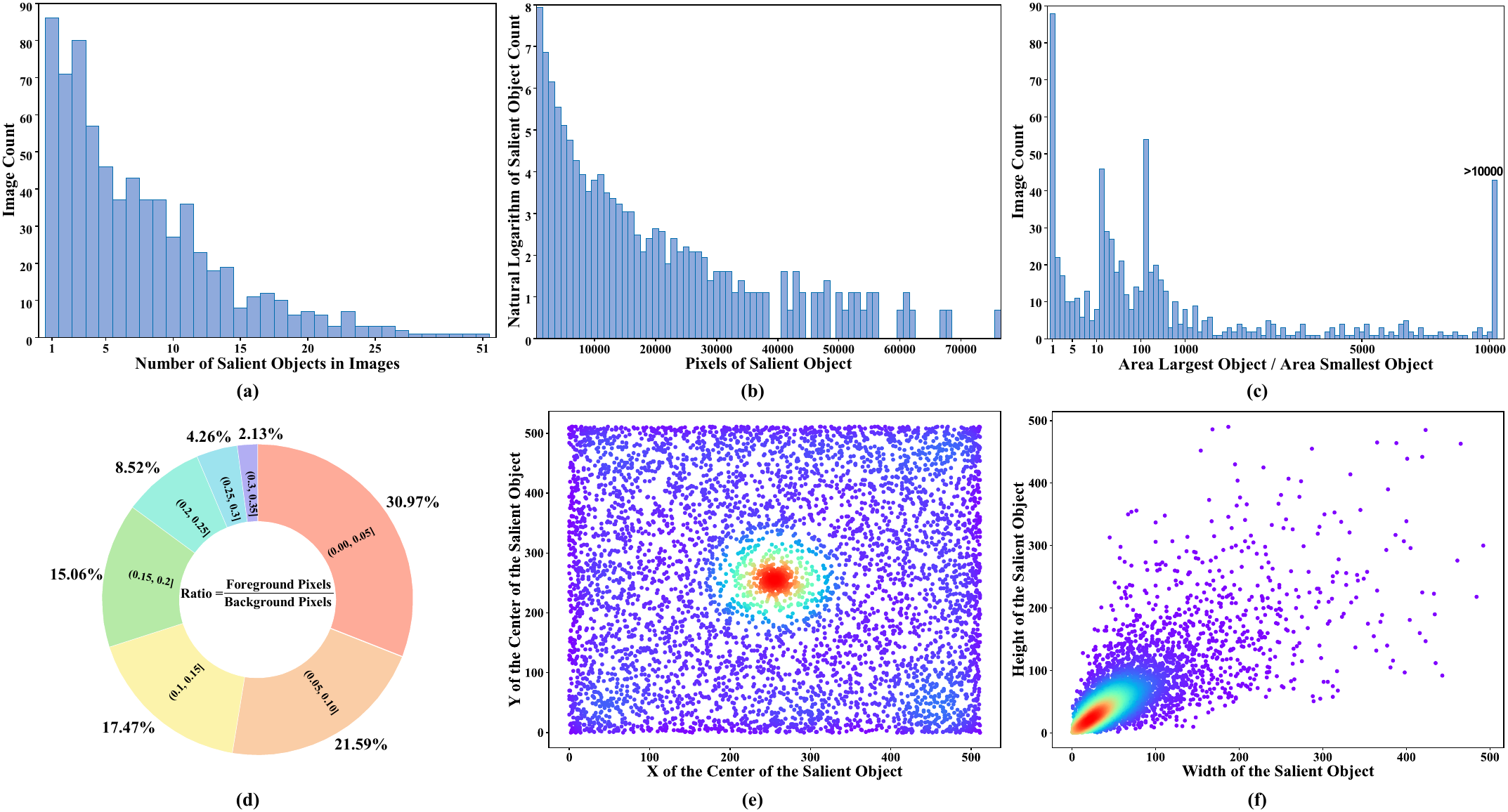}
  \caption{Statistical Overview of the HRSSD Dataset: \textbf{(a)} Distribution of salient object count per image. \textbf{(b)} Size distribution of salient objects. \textbf{(c)} Area ratio between largest and smallest objects. \textbf{(d)} Image distribution by foreground-to-background pixel ratio. \textbf{(e)} Center bias in salient object placement. \textbf{(f)} Width and height biases in salient object sizes.
   }
  \label{statistics in dataset}
\end{figure*}

\noindent\textbf{Number of Salient Objects per Image.}
Among the 704 images in the dataset, 86 contain a single salient object, while the remaining images exhibit multiple salient objects, as depicted in \cref{statistics in dataset}~\red{(a)}. On average, each image contains $7.56$ salient objects, underscoring the dataset's complexity and diversity in representing saliency.

\noindent\textbf{Size of Salient Object.}
Salient objects in hyperspectral remote sensing imagery vary significantly in size, as depicted in \cref{statistics in dataset}~\red{(b)}. \cref{statistics in dataset}~\red{(c)} illustrates the area ratio between the largest and smallest objects within an image, which can exceed 10000 on a $512 \times 512$ image. Additionally, \cref{statistics in dataset}~\red{(d)} shows the ratio of foreground to background pixels, revealing that approximately half of the images have a foreground-to-background pixel ratio below 0.1. The wide range of object sizes and the sparse presence of salient object pixels introduce significant challenges for salient object detection.

\noindent\textbf{Size Bias and Center Bias.}
The width-to-height ratios of salient objects and their spatial distribution are analyzed in \cref{statistics in dataset}~\red{(e)} and \cref{statistics in dataset}~\red{(f)}, respectively. These visualizations highlight substantial variation in object dimensions and their widespread locations within images. Notably, there is a tendency for salient objects to be concentrated near the image center, consistent with common salient object detection datasets~\cite{Shi2016ECSSD, Yang2013DUT, Li2014PASCAL, Li2015HKU, Wang2017DUTS}. This center bias reflects the natural tendency for important objects to appear centrally in images, further aligning the dataset with standard benchmarks.

\begin{figure*}[tp]
    \centering
    \includegraphics[width=\linewidth]{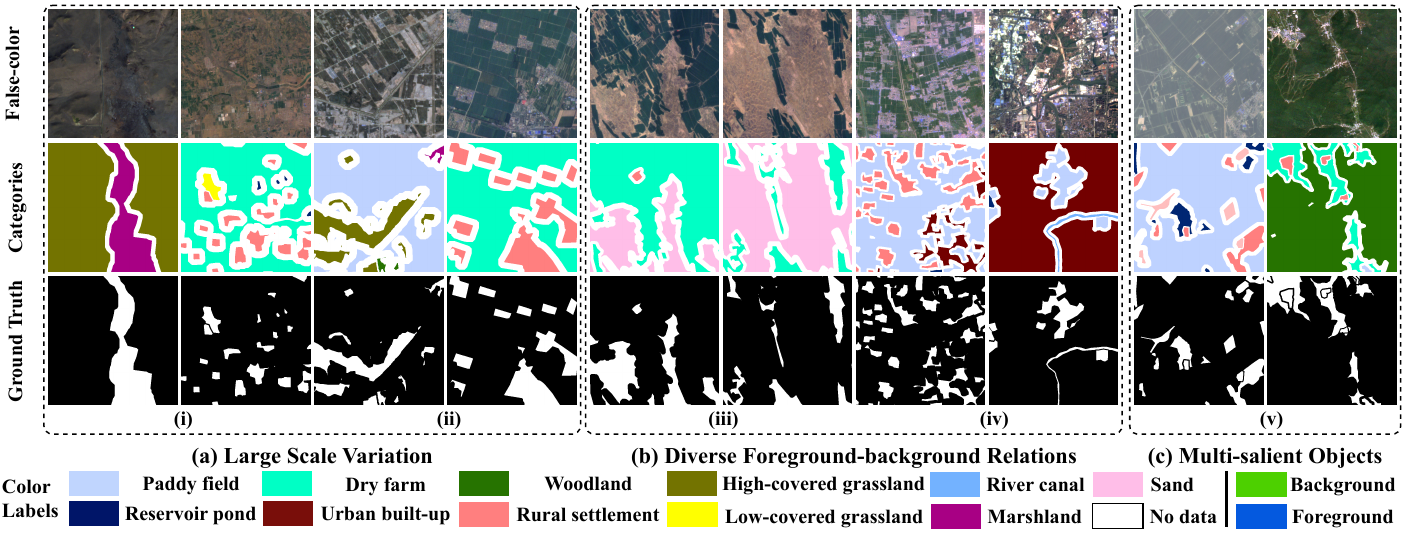}
    \caption{Challenges in HRSSD. (i) and (ii) illustrate \textbf{(a) large scale variations} present both across different images and within a single image. The complete reversal of foreground-background separation in (iii) and (iv) underscores the \textbf{(b) diverse foreground-background relations}. Many of the previous examples depict \textbf{(c) multiple salient objects}, with additional examples shown in (v).
}
    \label{challenge}
\end{figure*}

\subsection{Dataset Challenges}
\noindent\textbf{Large Scale Variation.} 
The HRSSD dataset exhibits significant variations in object sizes due to differences in land cover types and spatial scales, posing challenges in two key aspects. For instance, the scale disparity between marshland and reservoir ponds (\cref{challenge}~\red{(i)}) demonstrates cross-image scale variation, which occurs not only across different land cover types but also within the same type. This is further illustrated by high-coverage grassland and rural settlements (\cref{challenge}~\red{(ii)}), which occupy varying spatial proportions within images. Statistical analyses in \cref{statistics in dataset}~\red{(b)} and \cref{statistics in dataset}~\red{(c)} confirm substantial scale variations both across and within images. These variations challenge salient object detection methods to robustly capture the overall contours of large-scale objects while maintaining precision in detailing small-scale ones.

\noindent\textbf{Diverse Foreground-background Relations.}
The spatial distribution and spectral characteristics of land cover, combined with its contextual variability, lead to dynamic foreground-background relationships across environments. For example, in \cref{challenge}~\red{(iii)}, river canals and surrounding marshlands switch roles as foreground and background in different scenes. Similarly, urban built-up areas and adjacent farmland exhibit reversed foreground-background relationships in \cref{challenge}~\red{(iv)}. This variability requires saliency detection algorithms to effectively distinguish foreground from background in diverse and complex scenarios, adding another layer of difficulty.

\begin{figure*}[htp]
    \centering
    \includegraphics[width=\linewidth]{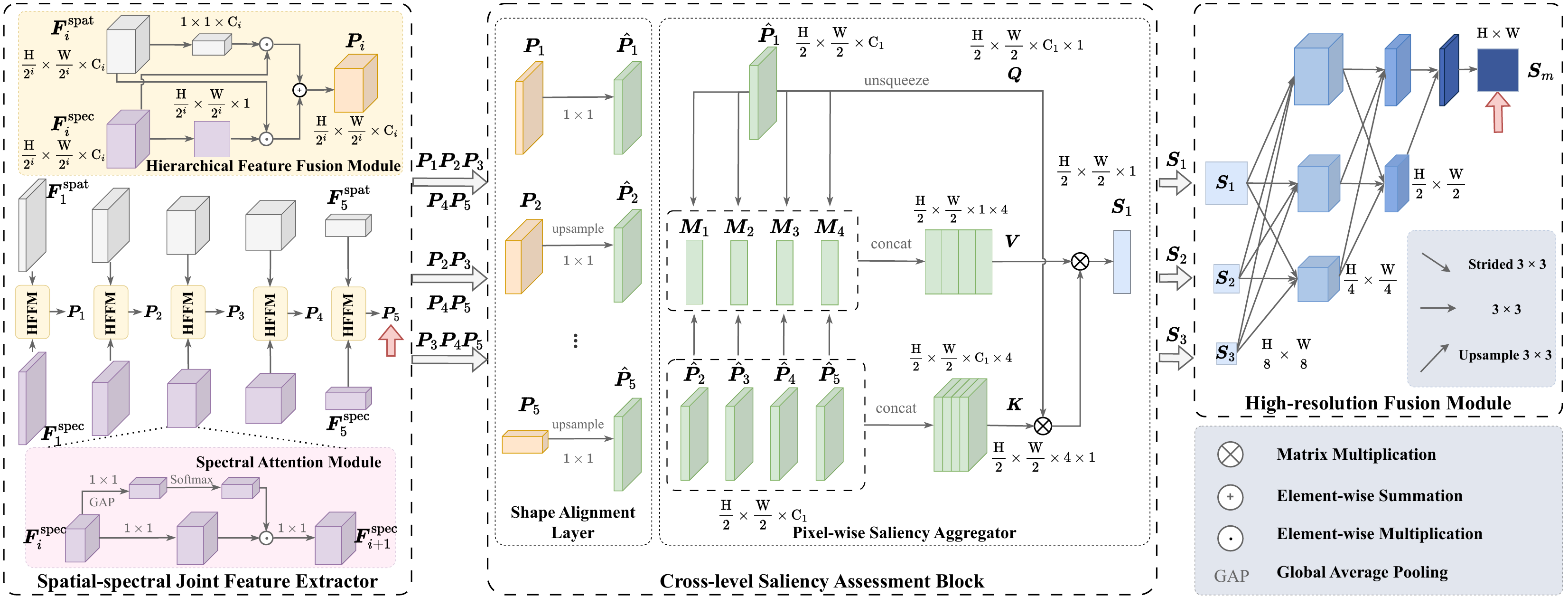}
        \caption{Overview of the Deep Spectral Saliency Network. The Spatial-spectral Joint Feature Extractor employs parallel branches to extract multi-level spatial-spectral features. The Cross-level Saliency Assessment Block computes pixel-wise attention-based similarities across levels to generate multi-scale similarity maps, which are then fused in the High-resolution Fusion Module. The structure of the Cross-level Saliency Assessment Block is illustrated with all features used as input. Red arrows indicate supervision signals.}    
    \label{Fig: overall architecture}
\end{figure*}

\noindent \textbf{Multi-salient Objects.}
As shown in \cref{challenge}~\red{(v)}, over $90\%$ of the images in HRSSD contain multiple salient objects, which may be closely positioned or exhibit diverse shapes. This complexity challenges the identification and segmentation of multiple salient objects within a single image.

\section{Deep Spectral Saliency Network}
Given a hyperspectral image $\boldsymbol{I} \in \mathbb{R}^{\rm H \times W \times C}$, the goal of salient object detection is to generate a saliency map $\boldsymbol{S}_\text{m} \in \mathbb{R}^{\rm H \times W \times 1}$ that highlights spectrally prominent objects or regions. This process can be expressed as:
\begin{equation}
\boldsymbol{S}_\text{m} = \boldsymbol{\Phi}(\boldsymbol{I}),
\end{equation}
where $\boldsymbol{\Phi}(\cdot)$ denotes the mapping function that transforms the input hyperspectral image into a saliency map. Our proposed Deep Spectral Saliency Network (DSSN) implements this mapping function. As depicted in~\cref{Fig: overall architecture}, DSSN begins with the Spatial-spectral Joint Feature Extractor, which employs parallel branches to extract multi-level spatial and spectral features. Subsequently, the Cross-level Saliency Assessment Blocks compute pixel-wise attention-based similarities across levels to produce multi-scale similarity maps. Finally, the High-resolution Fusion Module integrates these maps, enabling cross-resolution information exchange and generating the final saliency map.

\subsection{Spatial-spectral Joint Feature Extractor}
To integrate spatial-spectral information inherent in HSIs and enhance feature representation, we propose a Spatial-spectral Joint Feature Extractor (SJFE). As shown in~\cref{Fig: overall architecture}, SJFE consists of two parallel branches (Spatial Branch and Spectral Branch) for multi-level feature extraction, followed by a Hierarchical Feature Fusion module to generate integrated spatial-spectral representations.

\noindent \textbf{Spatial Branch} employs an off-the-shelf backbone network to extract hierarchical spatial features $\boldsymbol{F}^\text{spat} = \{\boldsymbol{F}^\text{spat}_i\}_{i=1}^5$. Each level $i$ produces features $\boldsymbol{F}^\text{spat}_i \in \mathbb{R}^{\frac{\mathrm{H}}{2^i} \times \frac{\mathrm{W}}{2^i} \times \mathrm{C}_i}$, where $\rm H$ and $\rm W$ denote input dimensions, and $\mathrm{C}_i$ represents channel count at scale $i$.

\noindent \textbf{Spectral Branch} integrates Spectral Attention Modules to extract discriminative spectral features $\boldsymbol{F}^\text{spec} = \{\boldsymbol{F}^\text{spec}_i\}_{i=1}^5$ with dimension alignment to the spatial branch. It first generates channel-wise attention weights $\boldsymbol{v} \in \mathbb{R}^{\mathrm{C}_i}$ through:
\begin{equation}
   \boldsymbol{v} = \sigma \left ( \boldsymbol{f}_\text{GAP} \left ( \boldsymbol{f}_\text{C1} \left ( \boldsymbol{I} \right ) \right ) \right ).
\end{equation} 
The functions $\boldsymbol{f}_\text{C1}(\cdot)$, $\boldsymbol{f}_\text{GAP}(\cdot)$, and $\sigma(\cdot)$ represent a $1 \times 1$ convolution, global average pooling operation, and the Softmax activation, respectively. Features are then refined via:
\begin{equation}
    \boldsymbol{F}^\text{spec}_1 = \boldsymbol{f}_\text{C3} \left ( \boldsymbol{v} \odot \left ( \boldsymbol{f}_\text{C2} \left (\boldsymbol{I} \right ) \right ) \right ).
\end{equation}
In this equation, $\boldsymbol{f}_\text{C2}(\cdot)$ and $\boldsymbol{f}_\text{C3}(\cdot)$ are $1 \times 1$ convolutions with distinct parameters, while $\odot$ indicates element-wise multiplication with broadcasting.

\noindent \textbf{Hierarchical Feature Fusion.} 
Cross-modal interaction is achieved through dual modulation at each pyramid level. Specifically, spatial features generate weight maps to recalibrate spectral features via:
\begin{equation}
    \hat{\boldsymbol{F}}_i^\text{spec} = \boldsymbol{f}_\text{C4} \left ( \boldsymbol{F}^\text{spat}_i \right ) \odot \boldsymbol{F}^\text{spec}_i,
\end{equation}
where $\boldsymbol{f}_\text{C4}(\cdot)$ denotes a $1 \times 1$ convolution for channel adjustment. Meanwhile, spectral features produce channel weights through enhance spatial features:
\begin{equation}
    \hat{\boldsymbol{F}}_i^\text{spat} = \boldsymbol{f}_\text{GAP} \left ( \boldsymbol{F}^\text{spec}_i \right ) \odot \boldsymbol{F}^\text{spat}_i.
\end{equation}
Final fusion combines modulated features through element-wise summation:
\begin{equation}
    \boldsymbol{P}_i = \hat{\boldsymbol{F}}_i^\text{spec} \oplus \hat{\boldsymbol{F}}_i^\text{spat},
\end{equation}
where $\oplus$ signifies element-wise summation. $\boldsymbol{P}_i \in \mathbb{R}^{\frac{\mathrm{H}}{2^i} \times \frac{\mathrm{W}}{2^i} \times \mathrm{C}_i}$ is the $i$-th layer of the spatial-spectral joint feature pyramid $\boldsymbol{P}=\{\boldsymbol{P}_i\}^5_{i = 1}$. These representations enable center-surround contrast analysis for saliency detection through subsequent multi-level feature comparison.

\subsection{Cross-level Saliency Assessment Block}
To conduct "center-surround spectral difference evaluation" and suppress erroneous responses in cluttered regions and emphasizes salient regions across scales, we introduce the Cross-Level Saliency Assessment Block (CSAB), the central component of our framework. As depicted in~\cref{Fig: overall architecture}, CSAB comprises two modules: the Shape Alignment Layer for dimension standardization and the Pixel-wise Saliency Aggregator for similarity evaluation and error suppression via pixel-wise attention mechanisms. 

The proposed Deep Spectral Saliency Network integrates three CSABs with shared architecture but varying input configurations. Each CSAB processes multi-level features to generate intermediate saliency maps $\boldsymbol{S} = \{\boldsymbol{S}_i\}_{i=1}^3$ at resolutions $\frac{\rm H}{2^i} \times \frac{\rm W}{2^i} \times 1$, where finer maps capture local details and coarser maps enhance object coverage.

\noindent \textbf{Shape Alignment Layer.} 
This layer aligns spatial and channel dimensions of input features through interpolation and linear transformation: 
\begin{equation}
    \hat{\boldsymbol{P}}_{i} = \boldsymbol{f}_{\text{up}} \left ( \boldsymbol{W}_i(\boldsymbol{P}_i) \right ), \quad i=1,2,3,4,5,
\end{equation}
where $\boldsymbol{W}_i$ denotes $1{\times}1$ convolution for channel adjustment and $\boldsymbol{f}_{\text{up}}(\cdot)$ upsamples features to a unified spatial resolution of $\frac{\rm H}{2} \times \frac{\rm W}{2}$ with preserved channel count $\rm C_1$.

\noindent \textbf{Pixel-wise Saliency Aggregator.}
Following Liang~\etal~\cite{Liang2013SODHS}, center-surround analysis is conducted by computing Euclidean distances between the highest-level feature $\hat{\boldsymbol{P}}_1$ and lower-level features $\hat{\boldsymbol{P}}_{1+j}$:
\begin{equation}
    \boldsymbol{M}_j = \left \| \hat{\boldsymbol{P}}_1 - \hat{\boldsymbol{P}}_{1+j} \right \|, \quad j=1,2,3,4.
\end{equation}
The obtained similarity maps $\boldsymbol{M}_j$ shares resolutions at $\frac{\rm H}{2} \times \frac{\rm W}{2} \times 1$.
To suppress erroneous responses, a pixel-level attention mechanism adaptively weights similarity maps. For each pixel $x$, the top-level feature $\hat{\boldsymbol{P}}_1$ serves as the \textit{query} $\boldsymbol{Q}(x) \in \mathbb{R}^{1{\times}1{\times}C_1}$, while lower-level features $\hat{\boldsymbol{P}}_{1+j}$ provide \textit{keys} $\boldsymbol{K}_j(x) \in \mathbb{R}^{1{\times}1{\times}C_1}$. Importance weights $\boldsymbol{A}(x) \in \mathbb{R}^{1{\times}1{\times}4}$ are computed as:
\begin{equation}
    \boldsymbol{A}(x) = \frac{\exp \left ( \boldsymbol{Q}(x)^\top \boldsymbol{K}_j(x) \right )}{\sum_{\forall j} \exp \left ( \boldsymbol{Q}(x)^\top \boldsymbol{K}_j(x) \right )}.
\end{equation}
Such weights, serving as \textit{value}, are applied to aggregate similarity maps into the final saliency map:
\begin{equation}
    \boldsymbol{S}_1(x) = \sum_{j=1}^4 \boldsymbol{A}(x) \otimes \boldsymbol{V}_j(x),
\end{equation}
where $\otimes$ denotes matrix multiplication. The implementation details are provided in~\cref{pserudo-code}. 

\begin{algorithm}[tp]
\centering
\renewcommand{\arraystretch}{0.5}
\caption{Pytorch-like Pseudo Code of CSAB.} 
\label{pserudo-code}
\begin{lstlisting}[language={Python}]
# feature_pyramid: input tensor
# q_index: a layer index
# k_indexes: list of layer indexes of the pyramid
# q_proj: a linear projection layer
# k_proj, v_proj: lists of linear projection layer
# dim: hidden_dim in CSAB

B, _, H, W = feature_pyramid[q_index].shape
Q = q_proj(feature_pyramid[q_index])  
Ks, Vs = [], []
for i in range(0, len(k_indexes)):
    K = interpolate(feature_pyramid[k_indexes[i]],(H,W))
    K = k_proj[i](rearrange(K,'b c h w -> b (h w) c'))
    Ks.append(K)

    V = interpolate(feature_pyramid[k_indexes[i]],(H,W))
    V = v_proj[i](rearrange(V,'b c h w -> b (h w) c'))
    Vs.append(pairwise_distance(Q,V,p=2).unsqueeze(-1))
    
Q = Q.view(B, -1, 1, dim)
K = cat(Ks, dim=-1).view(B, -1, len(k_indexes), dim)
V = cat(Vs, dim=-1).unsqueeze(-1)

attention_scores = matmul(Q, K.transpose(-1, -2))
attention_probs = softmax(attention_scores, dim=-1)
smap = matmul(attention_probs, V)
return rearrange(smap,'b (h w) 1 c -> b c h w', h=H)
\end{lstlisting}
\end{algorithm}

\subsection{High-resolution Fusion Module}
As emphasized in~\cite{Wang2021Deep,MCAT}, the fusion of multi-scale features is crucial for detecting objects of varying sizes. To this end, we propose the High-Resolution Fusion Module (HRFM), which employs a bottom-up fusion strategy to progressively aggregate information from high-resolution ($\boldsymbol{S}_1$) to low-resolution ($\boldsymbol{S}_3$) saliency maps, generating a final prediction with refined details. As shown in~\cref{Fig: overall architecture}, HRFM consists of three cascaded fusion stages with decreasing channel counts.

Specifically, the first layer accepts three intermediate saliency maps $\left \{ \boldsymbol{S}_l \right \}^3_{l=1}$ and generates larger-sized features $\left \{ \boldsymbol{F}_r^\text{o} \right \}^3_{r=1}$:
\begin{equation}
    \boldsymbol{F}^\text{o}_{r} = \sum_{l=1}^{3}\boldsymbol{f}_{lr}(\boldsymbol{S}_l),
\end{equation}
where $\boldsymbol{f}_{lr}(\cdot)$ denotes resolution-specific transformations, dependent on the input index $l$ and the output index $r$. When $l \ge r$, $\boldsymbol{f}_{lr}(\cdot)$ is implemented by bilinear interpolation followed by a convolution layer; when $l = r-1$, $\boldsymbol{f}_{lr}(\cdot)$ consists of a single convolution layer; when $l = r-2$, $\boldsymbol{f}_{lr}(\cdot)$ includes a convolution with a stride of 2. All convolution operations utilize $3 \times 3$ kernels, which extracts the spatial features more effectively, compared to $1 \times 1$ kernels.

Subsequent stages follow similar fusion patterns with reduced channel dimensions to balance computational efficiency, inversely mirroring the feature pyramid structure in the SJFE. This design ensures effective detection of multiple salient objects of varying sizes, even when they are closely spaced.

\begin{table*}[htp]
    \centering
    \caption{Quantitative results on our HRSSD dataset and efficiency analysis.}
    \label{Tab: quantitative}
    \setlength{\tabcolsep}{4.8mm}
    \begin{tabular}{l|c|cccc|ccc}
    \toprule[1.2pt]
        Method & Year & MAE $\downarrow$ & $F_\beta$ $\uparrow$ & $E_\xi$ $\uparrow$ & $S_\alpha$ $\uparrow$ & FLOPs (G) & \#Params (M) & Speed (FPS) \\ 
        \midrule
        \multicolumn{9}{c}{\textit{Methods for \textbf{Optical Remote Sensing Images}}} \\
        \midrule
        MJRBM~\cite{Tu2022ORSI4199} & 2022 & 0.078  & 0.497  & 0.764  & 0.643  & 95.81 & 43.54 & 40.09 \\
        FSMINet~\cite{Shen2022FSMINet} & 2022 & 0.073  & 0.387  & 0.664  & 0.578  & 11.83 & 3.56 & 40.16  \\
        CorrNet~\cite{Li2022CorrNet}& 2022 &  0.067  & 0.422  & 0.676  & 0.573  & 21.31 & 4.07 & 88.07  \\
        ACCoNet~\cite{Li2023ACCoNet} & 2023 & 0.064  & 0.487  & 0.722  & 0.629  & 184.50 & 102.55 & 55.12  \\
        SeaNet~\cite{Li2023SeaNet} & 2023 & 0.075  & 0.408  & 0.713  & 0.580 & 1.81 & 2.75 & 76.28  \\ 
        MEANet~\cite{Liang2024MEANet} & 2024 & 0.101 & 0.471 & 0.723 & 0.622 & 11.34 & 3.27 & 31.85 \\
        \midrule
        \multicolumn{9}{c}{\textit{Methods for \textbf{Optical Natural Scene Images}}} \\
        \midrule
        CTDNet~\cite{Zhao2021CTDNet} & 2021 & 0.069  & 0.442  & 0.719  & 0.605  & 6.15 & 11.83 & 140.97 \\
        TRACER~\cite{lee2022TRACER} & 2022 & 0.066  & 0.466  & 0.735  & 0.621  & 2.17 & 3.90 & 13.37  \\
        BBRF~\cite{Ma2023BBRF} & 2023 & 0.066  & 0.469  & 0.696  & 0.608 & 46.41 & 74.01 & 27.25  \\
        MENet~\cite{Wang2023MENet} & 2023 & \textbf{0.063} & 0.477 & 0.729 & 0.616 & 94.66 & 27.83 & 15.37 \\
        ADMNet~\cite{ADMNet} & 2024 & 0.077 & 0.337 & 0.635 & 0.539 & \textbf{0.87} & 0.84 & 20.87 \\
        \midrule
        \multicolumn{9}{c}{\textit{Methods for \textbf{Hyperspectral Natural Scene Images}}} \\
        \midrule
        SED~\cite{Liang2013SODHS} & 2013 & 0.108 & 0.281 & 0.621 & 0.518 & - & - & 2.24 \\
        SG~\cite{Liang2013SODHS} & 2013 & 0.120 & 0.262 & 0.644 & 0.510 & - & - & 2.24 \\
        SUDF~\cite{imamouglu2019salient} & 2019 & 0.221 & 0.148 & 0.479 & 0.449 & 82.90 & \textbf{0.10} & 0.51 \\ 
        SMN~\cite{SMN} & 2024 & 0.084 & 0.331 & 0.652 & 0.546 & 14.58 & 7.27 & 35.91 \\ 
        \midrule
        \multicolumn{9}{c}{\textit{Methods for \textbf{Hyperspectral Remote Sensing Images}}} \\
        \midrule
        CSCN~\cite{CSCN} & 2024 & 0.069 & 0.526 & 0.767 & 0.644 & 286.50 & 7.38 & 23.78 \\
        SAHRNet~\cite{SAHRNet} & 2024 & 0.095 & 0.442 & 0.677 & 0.599 & 692.04 & 2.65 & 170.54 \\
        MambaHSI~\cite{MambaHSI} & 2024 & 0.080 & 0.488 & 0.749 & 0.636 & 18.95 & 0.556 & 206.12 \\
        MambaLG~\cite{MambaLG} & 2025 & 0.075 & 0.491 & 0.724 & 0.630 & 94.65 & 1.55 & \textbf{206.62} \\
        DSTC~\cite{DSTC} & 2025 & 0.076 & 0.490 & 0.754 & 0.603 & 17.25 & 4.13 & 4.79 \\
        \rowcolor{Gray} 
        DSSN (Ours) & 2025 & 0.069 & \textbf{0.556} & \textbf{0.769} & \textbf{0.655} & 4.95 & 5.15 & 43.12 \\
    \bottomrule[1.2pt]
    \end{tabular}
\end{table*}

\subsection{Objective Function}
To optimize feature learning and saliency prediction, we employ dual supervision using standard loss functions applied to the final feature map $\boldsymbol{P}_5$ from the Spatial-spectral Joint Feature Extractor, and the ultimate saliency map $\boldsymbol{S}_\text{m}$. The training objective $\mathcal{L}$ combines two widely-used loss functions $\mathcal{L}_\text{BCE}$ and $\mathcal{L}_\text{IoU}$:
\begin{equation}
    \begin{aligned}
    \mathcal{L} &= \mathcal{L}_\text{BCE}(\boldsymbol{P}_5^\prime, \boldsymbol{G}) + \mathcal{L}_\text{IoU}(\boldsymbol{P}_5^\prime, \boldsymbol{G}) 
    \\ &+ \mathcal{L}_\text{BCE}(\boldsymbol{S}_\text{m}, \boldsymbol{G}) + \mathcal{L}_\text{IoU}(\boldsymbol{S}_\text{m}, \boldsymbol{G}),
    \end{aligned}
\end{equation}
where $\boldsymbol{P}_5^\prime$ is the final feature map $\boldsymbol{P}_5$ after channel alignment with the ground-truth saliency map $\boldsymbol{G}$.
Binary Cross Entropy (BCE) Loss $\mathcal{L}_{\text{BCE}}$ enforces pixel-wise accuracy:
\begin{equation}
    \mathcal{L}_{\text{BCE}}(\boldsymbol{X},\boldsymbol{G})=-\sum {\left [ \boldsymbol{G} \log(\boldsymbol{X})+(1-\boldsymbol{G}) \log(1-\boldsymbol{X}) \right ]}.
\end{equation}
Meanwhile, Intersection over Union (IoU) Loss $\mathcal{L}_{\text{IoU}}$ enhances shape consistency:
\begin{equation} \small
    \mathcal{L}_{\text{IoU}}(\boldsymbol{X}, \boldsymbol{G}) = 1 - \frac{\sum\limits_{r=1}^{\mathrm{H}}\sum\limits_{c=1}^{\mathrm{W}} \boldsymbol{X}(r,c)\boldsymbol{G}(r,c)}{\sum\limits_{r=1}^{\mathrm{H}}\sum\limits_{c=1}^{\mathrm{W}} [\boldsymbol{X}(r,c)+\boldsymbol{G}(r,c)-\boldsymbol{X}(r,c)\boldsymbol{G}(r,c)]},
\end{equation}
where $\boldsymbol{X} \in \{\boldsymbol{P}_5^\prime, \boldsymbol{S}_\text{m}\}$ denotes the network predictions, and $\mathrm{H}$ and $\mathrm{W}$ represent the height and width, respectively.

\section{Experiments}
We evaluate the performance of multiple methods using our HRSSD dataset and assess the efficacy of our DSSN on hyperspectral natural scene salient object detection datasets, HSOD-BIT~\cite{qin2024dmssn} and HS-SOD~\cite{Imamoglu2018HSSOD}.

\subsection{Experimental Settings}
\noindent \textbf{Implementation Details.}  
The proposed model is implemented on a single NVIDIA RTX 3090 GPU with an Intel Xeon Gold 5218R CPU. In the Spatial-spectral Joint Feature Extractor, MobileNetV2~\cite{Sandler_2018_CVPR} serves as the backbone for the spatial branch, while the spectral branch integrates our proposed Spectral Attention Module as a replacement for MobileNetV2's Inverted Residual Blocks. Hyperspectral images are uniformly resampled to a spatial resolution of $256 \times 256$ pixels. To prevent overfitting, data augmentation techniques including random horizontal flipping and cropping are applied. Following the protocol in~\cite{Li2022WHUOHS}, input values are scaled by a factor of \(1/10000\) to preserve original radiometric characteristics instead of standard normalization. The model is optimized using the Nadam optimizer with an initial learning rate of \(3 \times 10^{-3}\), which is dynamically adjusted via a cosine annealing schedule. Training is performed for 100 epochs with a batch size of 16 samples.

\noindent \textbf{Competing Methods.}  
As the first to propose a salient object detection method specifically tailored for hyperspectral remote sensing images, we have adapted recent hyperspectral image classification approaches to facilitate a comprehensive comparison. These include CSCN~\cite{CSCN}, SAHRNet~\cite{SAHRNet}, MambaHSI~\cite{MambaHSI}, MambaLG~\cite{MambaLG}, and DSTC~\cite{DSTC}. To further enrich our analysis, we incorporate three methods designed explicitly for hyperspectral natural scene images: SED~\cite{Liang2013SODHS}, SG~\cite{Liang2013SODHS}, SUDF~\cite{imamouglu2019salient}, and the most recent open-source state-of-the-art approach, SMN~\cite{SMN}.

\begin{figure*}[ht]
  \centering
  \includegraphics[width=\linewidth]{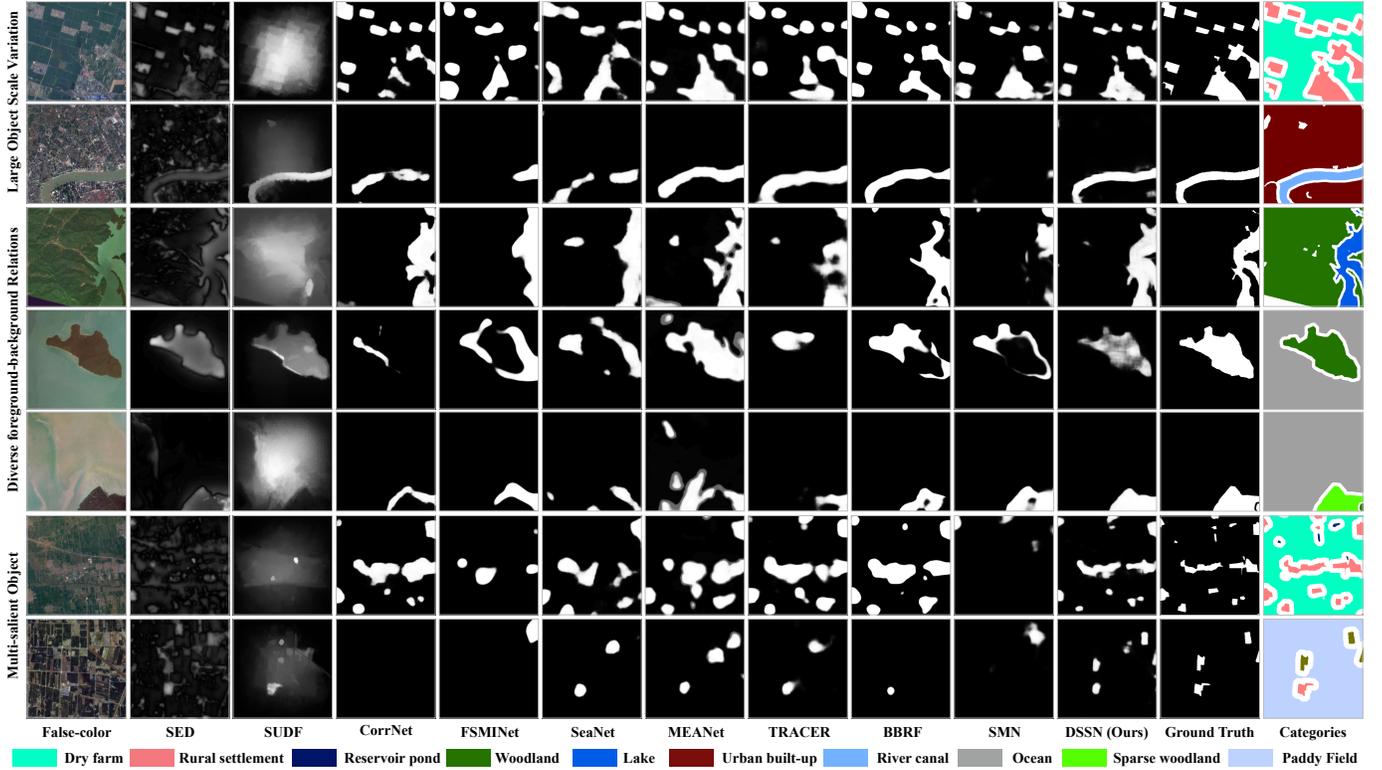}
  \caption{Visualization results on the HRSSD dataset demonstrate the efficacy of our proposed DSSN. When confronted with the multiple challenges inherent in the dataset, our DSSN exhibits a substantial detection advantage over comparative methods.}
  \label{Fig: qualitative HR-SOD}
\end{figure*}

To broaden the scope of our evaluation, we also include seven methods originally developed for optical remote sensing images: MJRBM~\cite{Tu2022ORSI4199}, FSMINet~\cite{Shen2022FSMINet}, CorrNet~\cite{Li2022CorrNet}, ACCoNet~\cite{Li2023ACCoNet}, SeaNet~\cite{Li2023SeaNet}, and MEANet~\cite{Liang2024MEANet}. Additionally, we assess several methods designed for optical natural scene images, such as CTDNet~\cite{Zhao2021CTDNet}, TRACER~\cite{lee2022TRACER}, BBRF~\cite{Ma2023BBRF}, MENet~\cite{Wang2023MENet}, and ADMNet~\cite{ADMNet}. Methods designed for RGB images accept false-color images as input. All competing methods are implemented using their default settings.

\noindent \textbf{Evaluation Metrics.}  
Quantitative assessment employs four standardized metrics:  
1. \textit{Mean Absolute Error} (MAE) quantifies pixel-wise prediction accuracy through average absolute differences between saliency maps and ground truth.  
2. \textit{Maximum F-measure} ($F_\beta$) balances precision-recall trade-off via harmonic mean calculation.  
3. \textit{Enhanced-alignment Measure} ($E_\xi$)evaluates both local pixel matching and global distribution consistency.  
4. \textit{Structure Measure} ($S_\alpha$) emphasizes spatial coherence by assessing object-level structural similarity.

\subsection{Experiment on HRSSD}
\noindent\textbf{Quantitaive Results.}
As demonstrated in~\cref{Tab: quantitative}, our DSSN achieves state-of-the-art performance across multiple evaluation metrics on the HRSSD dataset. Specifically, DSSN outperforms competing methodologies in terms of $F_\beta$, $E_\xi$, and $S_\alpha$. Although MENet achieves a marginally lower MAE score, its overall performance is limited by suboptimal scores in other metrics, particularly $F_\beta$. The inferior performance of methods designed for optical images underscores the critical importance of incorporating spectral information. 

Methods tailored for hyperspectral natural scene images, such as SUDF and SMN, also exhibit limited effectiveness. While these methods leverage spectral information to some extent, they fail to capture the essence of "center-surround spectral difference evaluation," which is pivotal for addressing challenges such as large scale variation, diverse foreground-background relations, and multi-salient objects. 

Despite significant progress in classification tasks, current hyperspectral image classification methods show limited adaptability to salient object detection. This limitation underscores the need for specialized algorithms and dedicated datasets specifically designed for HRSI-SOD.

\noindent\textbf{Efficiency Analysis.}
In addition to superior detection performance, DSSN achieves an optimal balance between computational efficiency and accuracy. As shown in \cref{Tab: quantitative}, DSSN requires 4.95 G FLOPs and 5.15 million parameters, making it significantly more efficient than many competing methods. For example, ACCoNet and CSCN incur substantially higher computational costs (184.50 G and 286.50 G FLOPs, respectively), despite their relatively high MAE scores. Similarly, SUDF, while having the smallest parameter count (0.10 M), suffers from poor detection performance due to its reliance on minimal convolutional layers for feature extraction.
DSSN's efficiency is further demonstrated by its inference speed of 43.12 FPS, surpassing most deep learning-based methods. Notably, while MambaLG achieves the highest inference speed (206.62 FPS), its detection performance is suboptimal, with an MAE of 0.075 and an $F_\beta$ score of 0.491.

\begin{table*}[htp]
    \centering
    \caption{Quantitative results on HSOD-BIT and HS-SOD datasets.}
    \label{Tab: HSOD results}
    \setlength{\tabcolsep}{4.8mm}
    \begin{tabular}{l|l|cccc|cccc} 
        \toprule
        \multirow{2}{*}{Method} & \multirow{2}{*}{Year} & \multicolumn{4}{c|}{HSOD-BIT} & \multicolumn{4}{c}{HS-SOD}  \\ 
        \cmidrule{3-10} & & MAE $\downarrow$ & $S_\alpha$ $\uparrow$ & AUC $\uparrow$ & CC $\uparrow$       & MAE $\downarrow$ & $S_\alpha$ $\uparrow$ & AUC $\uparrow$ & CC $\uparrow$   \\ 
        \midrule
        SED~\cite{Liang2013SODHS} & 2013 & 0.130 & 0.500 & 0.753 & 0.303 & 0.130 & 0.466 & 0.793 & 0.201 \\
        SG~\cite{Liang2013SODHS}  & 2013 & 0.182 & 0.543  & 0.791  & 0.370 & 0.171  & 0.521  & 0.808  & 0.268   \\
        SUDF~\cite{imamouglu2019salient}  & 2019 & 0.150 & 0.685  & 0.918  & 0.671 & 0.242  & 0.498  & 0.723  & 0.250   \\
        CTDNet~\cite{Zhao2021CTDNet} & 2021 & 0.042 & 0.837  & 0.937  & 0.805 & 0.105  & 0.513  & 0.687  & 0.306   \\
        TRACER~\cite{lee2022TRACER} & 2022 & 0.039 & 0.862  & \textbf{0.970} & 0.846 & 0.157  & 0.595  & 0.868  & 0.465   \\
        BBRF~\cite{Ma2023BBRF} & 2023 & \textbf{0.033} & 0.868  & 0.932  & 0.845 & 0.090  & 0.663  & 0.781  & 0.520 \\
        ADMNet~\cite{ADMNet} & 2024 & 0.057 & 0.813 & 0.911 & 0.740 & 0.116 & 0.590 & 0.784 & 0.383\\
        SMN~\cite{SMN} & 2024 & 0.039 & 0.869  & 0.969  & \textbf{0.849} & 0.069  & \textbf{0.767}  & 0.903  & 0.684 \\ 
        \midrule
        \rowcolor{Gray}
        DSSN (Ours) & 2025  & 0.035 & \textbf{0.871} & \textbf{0.970} & 0.840 & \textbf{0.067} & 0.755 & \textbf{0.926} & \textbf{0.701}  \\
        \bottomrule
\end{tabular}
\end{table*}

\begin{figure}[tp]
  \centering
  \includegraphics[width=\linewidth]{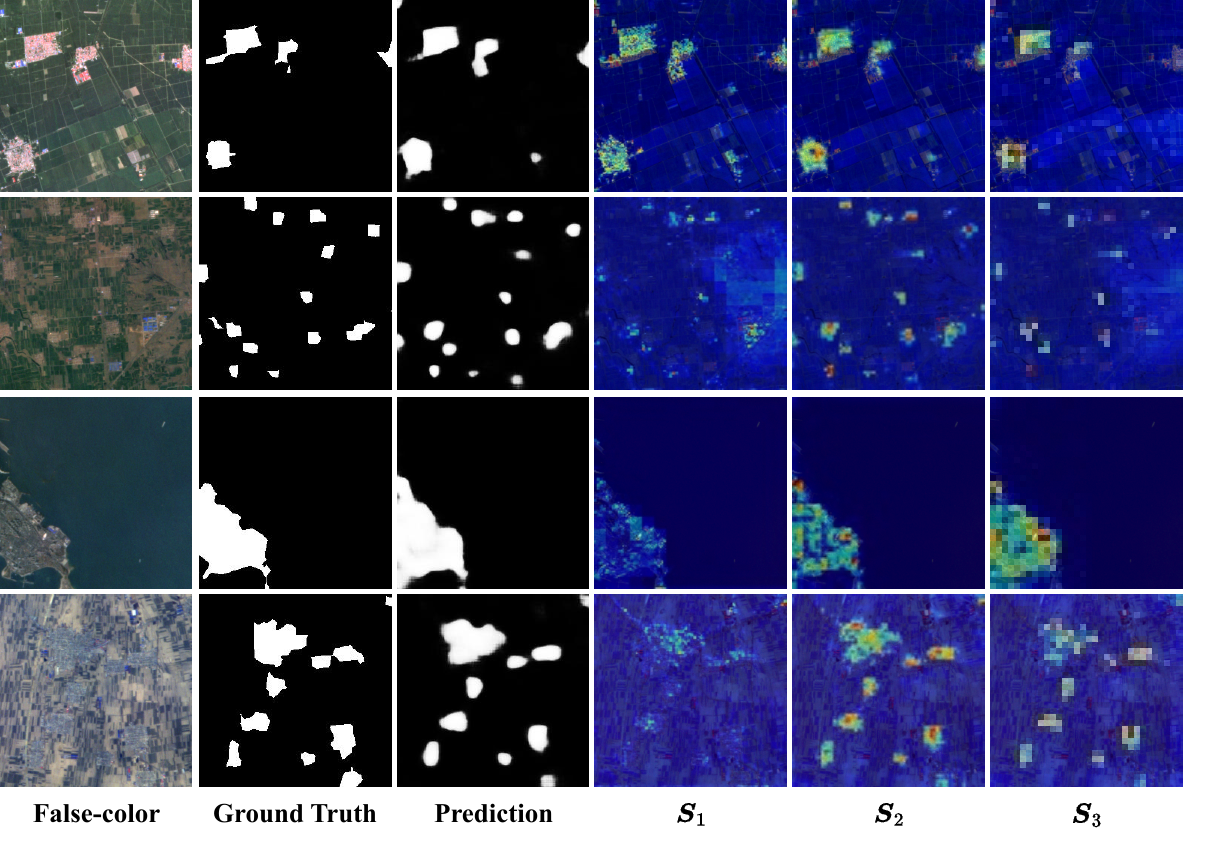}
  \caption{Visualization of intermediate saliency maps.}
  \label{Fig: saliency map vis}
\end{figure}

\noindent\textbf{Qualitative Results.}
As illustrated in~\cref{Fig: qualitative HR-SOD}, DSSN demonstrates significant advantages in handling complex challenges inherent in the HRSSD dataset:
\textbf{i) Large Scale Variation:} DSSN accurately detects both large-scale objects (\textit{e.g.}, rivers) and small-scale objects (\textit{e.g.}, rural settlements) with high precision. In contrast, methods like SMN and FSMINet struggle to comprehensively capture objects of varying scales.  
\textbf{ii) Diverse Foreground-Background Relations:} DSSN effectively identifies salient objects regardless of their roles as foreground or background. For instance, it accurately detects woodland as the foreground against an ocean background and vice versa, whereas methods like TRACER and SeaNet perform inconsistently across such scenarios. 
\textbf{iii) Multi-salient Objects:} DSSN comprehensively detects all small salient objects in images with multiple targets, whereas other methods often miss some objects. Additionally, DSSN excels in detecting small scattered objects (\textit{e.g.}, rural settlements) and complex-shaped objects (\textit{e.g.}, woodland islands).  

The superior performance of DSSN can be attributed to its innovative design components, specifically tailored to address the challenges in HRSI-SOD. The Spatial-spectral Joint Feature Extractor employs parallel spatial and spectral branches to enhance foreground-background discriminability by capturing complementary information from both domains. The Cross-Level Saliency Assessment Block adheres to the principle of "center-surround spectral difference evaluation" to generate initial similarity maps, while applying pixel-wise attention mechanisms to suppress erroneous responses in cluttered regions and emphasize salient regions across multiple scales. Additionally, the High-Resolution Fusion Module integrates a bottom-up fusion strategy with learned spatial upsampling to fully leverage the strengths of multi-scale saliency maps, recovering fine-grained details and enabling the detection of multiple salient objects of varying sizes. These innovations collectively make DSSN highly effective for hyperspectral remote sensing image salient object detection.

\noindent\textbf{Visualization of Intermediate Saliency Maps.}  
As shown in \cref{Fig: saliency map vis}, the saliency maps generated by the Cross-level Saliency Assessment Blocks reveal distinct characteristics at different resolutions. The high-resolution map $\boldsymbol{S}_1$ excels in capturing fine details of salient objects but is more prone to noise. In contrast, the lower-resolution map $\boldsymbol{S}_3$ demonstrates superior accuracy in holistic salient object detection, albeit with less precision in fine details.  

\subsection{Experiment on HSOD-BIT and HS-SOD}
\noindent\textbf{Datasets.}
The HSOD-BIT dataset contains 319 hyperspectral images of natural scenes, spanning a spectral range of 400–1000~nm with a spatial resolution of $1240 \times 1680$ pixels. It is divided into a training set of 255 images and a test set of 64 images. The dataset incorporates challenges such as similar foreground-background colors, overexposure, and uneven illumination to highlight the advantages of hyperspectral imaging under complex lighting conditions. The HS-SOD dataset comprises 60 hyperspectral images with a spectral range of 380–780~nm sampled at 5~nm intervals and a spatial resolution of $768 \times 1024$ pixels. It is split into a training set of 48 images and a test set of 12 images. Both datasets provide corresponding RGB images and ground truth binary masks.

\noindent \textbf{Implementation Details.}
For the HSOD-BIT dataset, we use a batch size of 48, while for HS-SOD, the batch size is set to 16. Training is conducted for 200 epochs across both datasets, with input images resized to $224 \times 224$ pixels.

\noindent \textbf{Competing Methods.}
We compare our method with state-of-the-art approaches designed for hyperspectral natural scene images, including SED~\cite{Liang2013SODHS}, SG~\cite{Liang2013SODHS}, SUDF~\cite{imamouglu2019salient}, and SMN~\cite{SMN}. Additionally, we evaluate it against methods developed for optical natural scene images, using false-color images generated from hyperspectral data as inputs. These include CTDNet~\cite{Zhao2021CTDNet}, TRACER~\cite{lee2022TRACER}, BBRF~\cite{Ma2023BBRF}, and ADMNet~\cite{ADMNet}. All competing methods are implemented using their default settings.

\begin{figure*}[htp]
  \centering
  \includegraphics[width=\linewidth]{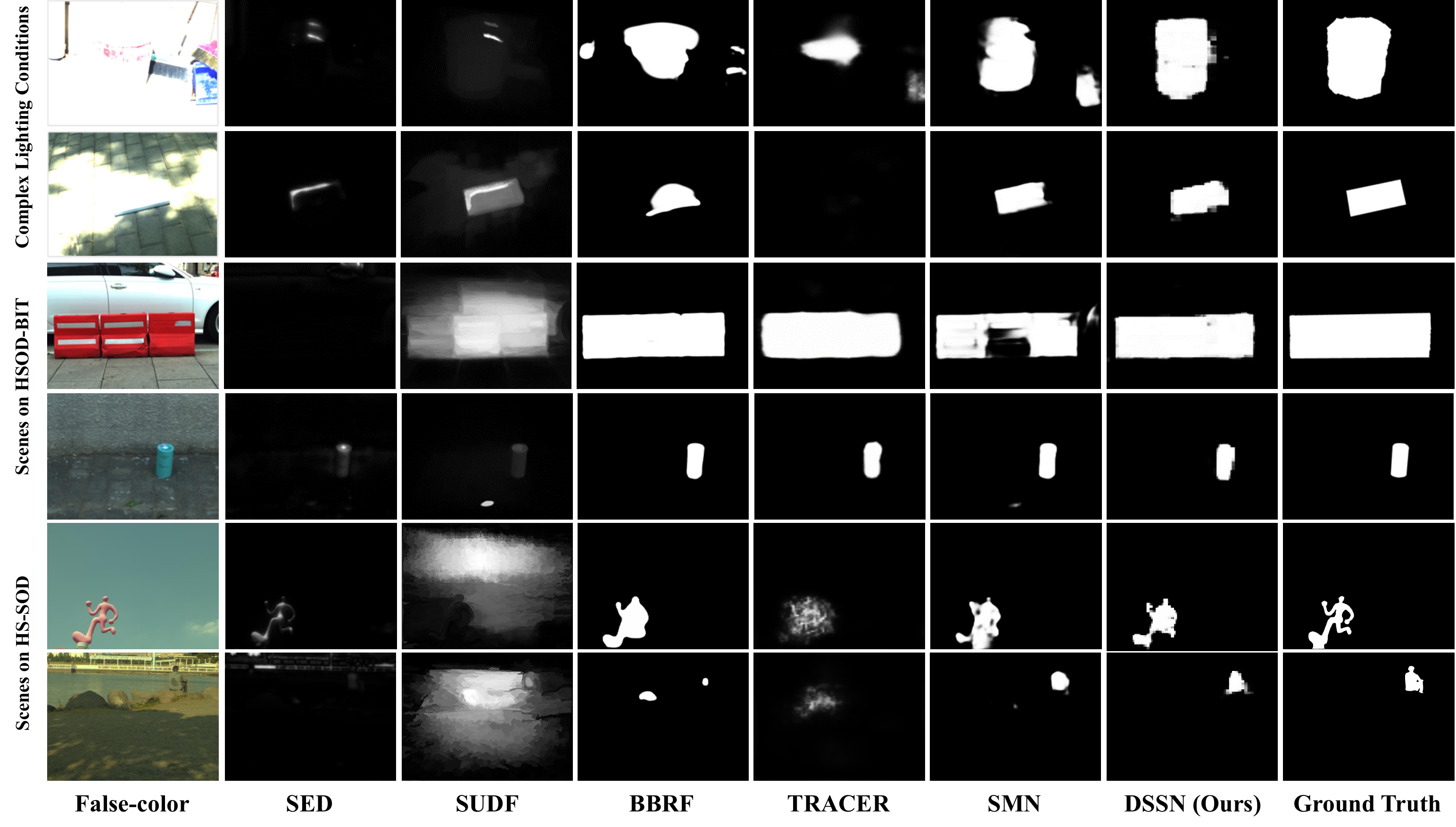}
  \caption{Visualization results on HSOD-BIT and HS-SOD datasets. DSSN demonstrates superior detection performance in various scenarios.}
  \label{Fig: HSOD vis}
\end{figure*}

\noindent \textbf{Evaluation Metrics.}
Performance is evaluated using four widely adopted metrics in HSOD tasks: Mean Absolute Error (MAE), Structure Measure ($S_\alpha$), Area Under the Curve (AUC), and Correlation Coefficient (CC). AUC reflects the model's ability to distinguish between salient and non-salient regions, while CC measures the alignment between predicted saliency maps and ground truth.

\noindent \textbf{Quantitative Results.}
As shown in \cref{Tab: HSOD results}, DSSN consistently achieves low MAE values and high $S_\alpha$, AUC, and CC scores across both datasets, demonstrating its effectiveness and robustness. On the HSOD-BIT dataset, DSSN achieves an MAE of $0.035$, outperforming all compared methods, with the closest competitor, BBRF, at $0.033$. DSSN attains an $S_\alpha$ of $0.871$, surpassing the best-performing baseline, SMN, which scores $0.869$. It matches TRACER for the highest AUC at $0.970$ and achieves a CC of $0.840$, slightly below SMN's $0.849$, indicating strong detection accuracy.

On the HS-SOD dataset, DSSN demonstrates robust performance with an MAE of $0.067$, the lowest among all methods, outperforming SMN's second-best result of $0.069$. For $S_\alpha$, DSSN scores $0.755$, second only to SMN's $0.767$. DSSN achieves an AUC of $0.926$, the second-highest after SMN's $0.903$, and a CC of $0.701$, showcasing competitive performance that exceeds most other methods.

\noindent \textbf{Qualitative Results.}
As illustrated in \cref{Fig: HSOD vis}, DSSN exhibits superior performance in salient object detection, particularly under complex lighting conditions and challenging scenes. Visualizations reveal that many methods struggle to accurately capture object boundaries, often resulting in blurred or incomplete detections. In contrast, DSSN provides precise boundary definitions and enhances object localization, closely resembling the ground truth.

On the HSOD-BIT dataset, unlike methods such as SED and SUDF, which frequently highlight irrelevant background regions or fail to produce high-contrast object areas, DSSN demonstrates minimal artifacts and achieves high accuracy. On the HS-SOD dataset, methods like SED and BBRF often struggle with finer structural details, resulting in fragmented or incomplete object contours. DSSN, however, effectively captures the entire object structure, producing outputs that closely align with the ground truth. Overall, DSSN demonstrates exceptional performance and robustness in addressing complex lighting conditions and diverse scenes, underscoring its superior generalization capability for hyperspectral natural scene images.

\subsection{Failure Cases}
\begin{figure}
    \centering
    \includegraphics[width=1\linewidth]{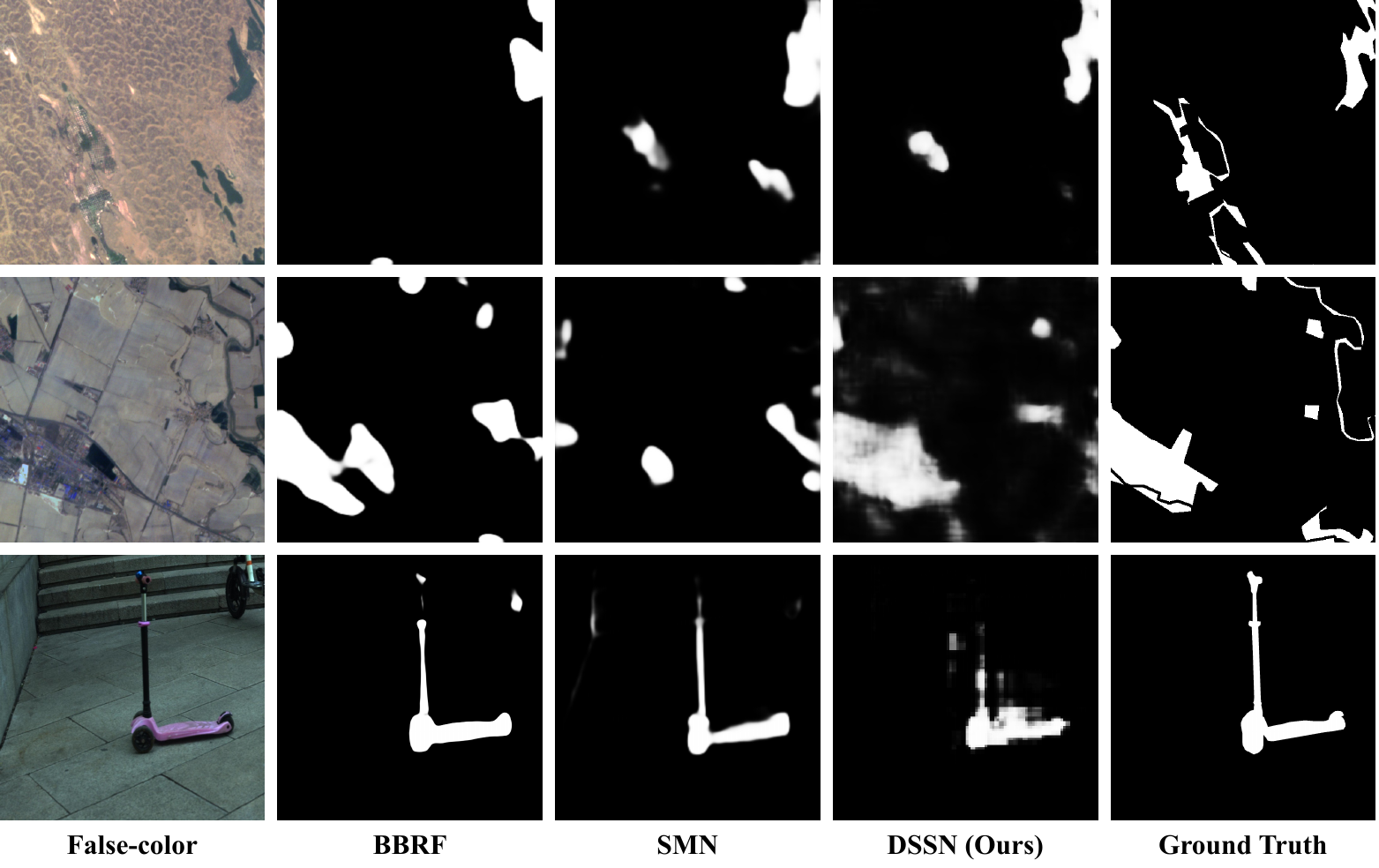}
    \caption{Illustration of failure cases. DSSN and other comparative methods encounter challenges when target regions exhibit thin connecting parts.}
    \label{fig: failure cases}
\end{figure}

As shown in \cref{fig: failure cases}, the proposed DSSN, along with other comparative methods, demonstrates limitations in handling target regions with thin connecting parts. For example, in the first two remote sensing scenes, these methods struggle to accurately detect finer details, such as small land areas or narrow water bodies. In the final natural scene, DSSN incompletely detects the scooter, capturing only its lower portion while failing to preserve the full structural details. Methods tailored for natural scenes, such as BBRF and SMN, achieve better performance.

\begin{table}[tp]
    \centering
    \small
    \caption{Ablation study of modalities in feature extraction.}
    \label{Tab: modality ablation}
    \setlength{\tabcolsep}{6mm}
    \renewcommand\arraystretch{1.26}
    \begin{tabular}{cc|cc}
        \toprule[1.2pt]
        Spatial & Spectral & $F_\beta^\text{max}$ $\uparrow$ & $S_\alpha$ $\uparrow$\\
        \midrule
        \have & \nohave & 0.500  & 0.635 \\
        \nohave & \have & 0.523  & 0.640 \\ 
        \have & \have & \textbf{0.556}  & \textbf{0.655}\\ 
        \bottomrule[1.2pt]
    \end{tabular}
\end{table}

\subsection{Ablation Study}
\noindent \textbf{Effect of input modalities.}
Within the Spatial-spectral Joint Feature Extractor, we evaluated the performance of extracting spatial and spectral features individually and in combination. The results in \cref{Tab: modality ablation} demonstrate that simultaneous extraction of spatial and spectral features significantly outperforms single-modality approaches. Specifically, spatial feature extraction alone shows reductions of $0.023$ and $0.005$ in key metrics compared to spectral feature extraction alone. This highlights the critical contribution of spectrum to HRSI-SOD.

\noindent \textbf{Effect of Cross-level Saliency Assessment Block.}
The effectiveness of the CSAB is demonstrated in the second row of \cref{Tab: ablation}. In this experiment, pyramid similarity maps were fused using direct summation instead of the proposed pixel-wise weighting scheme. By contrast, CSAB adaptively weights each similarity map, enabling more refined and context-aware fusion. This approach enhances the consolidation of salient features, leading to improved detection performance compared to the naive summation strategy.

\begin{table}[tp]
    \centering
    \small
    \caption{Ablation study of CSAB and HRFM.}
    \label{Tab: ablation}
    \setlength{\tabcolsep}{6mm}
    \begin{tabular}{cc|cc}
        \toprule[1.2pt]
        CSAB & HRFM & $F_\beta^\text{max}$ $\uparrow$ & $S_\alpha$ $\uparrow$\\
        \midrule
        \nohave &  \nohave & 0.268  & 0.481 \\ 
        \have &  \nohave & 0.506  & 0.636 \\
        \nohave &  \have & 0.537  & 0.644 \\ 
        \have &  \have & \textbf{0.556}  & \textbf{0.655}\\ 
        \bottomrule[1.2pt]
    \end{tabular}
\end{table}

\begin{table}[t]
    \centering
    \caption{Ablation study of loss functions.}
    \label{tab: loss ab}
    \setlength{\tabcolsep}{5mm}
    \begin{tabular}{ccc|cc} 
    \toprule[1.2pt]
    BCE & IoU & SSIM & $F_\beta^\text{max}$ $\uparrow$ & $S_\alpha$ $\uparrow$ \\ 
    \midrule
    \have & \nohave & \nohave & 0.503 & 0.623  \\
    \nohave & \have & \nohave & 0.508 & 0.615  \\
    \nohave & \nohave & \have & - & - \\
    \nohave & \have & \have & 0.508 & 0.607 \\
    \have & \nohave & \have & 0.448 & 0.602 \\
    \have & \have & \have & 0.528 & 0.635 \\
    \have & \have & \nohave & \textbf{0.556} & \textbf{0.655}  \\
    \bottomrule[1.2pt]
    \end{tabular}
\end{table}

\noindent \textbf{Effect of High-resolution Fusion Module.}
The third row of \cref{Tab: ablation} evaluates HRFM. Here, HRFM was replaced with a series of stacked convolutional layers, maintaining equivalent layer count and feature dimensionality. This alternative configuration achieved $F_\beta$ and $S_\alpha$ scores of $0.506$ and $0.636$, respectively, representing decreases of $0.050$ and $0.019$ compared to the original setup. These results underscore the importance of HRFM in preserving high-resolution features, facilitating inter-resolution information exchange, and effectively leveraging intermediate saliency maps for enhanced detection performance.

\noindent \textbf{Ablation of Loss Functions.}
To assess the impact of different loss functions, we included the Structural Similarity Index Measure (SSIM) Loss, which quantifies the structural similarity between predicted and ground truth maps. We performed an ablation study by removing each loss function individually to validate its effectiveness and examine the rationality of combining BCE Loss and IoU Loss. As shown in~\cref{tab: loss ab}, the combination of BCE Loss and IoU Loss achieves superior performance, while the removal of any component leads to performance degradation. Notably, when using only SSIM Loss, the model failed to converge, and no valid results were obtained.

\section{Future Work}
Future research efforts could collect hyperspectral images in a broader range of scenarios and explore advanced technologies such as data augmentation to increase the diversity and scale of datasets. Additionally, developing more sophisticated feature extraction methods capable of capturing fine-grained structural details will be critical for improving detection accuracy. Optimizing multi-scale fusion strategies remains essential to preserve intricate details while ensuring precise identification of salient regions in complex scenes. Furthermore, exploring the practical applicability of these advancements in domains such as military defense, mineral mapping, and atmospheric monitoring ould further enhance the model's robustness, adaptability, and generalization capabilities. 

\section{Conclusion}
In this work, we introduce the Hyperspectral Remote Sensing Saliency Dataset (HRSSD), which includes 704 hyperspectral images with 5327 pixel-level annotated salient objects. The dataset captures diverse challenges such as large-scale variations, complex foreground-background relationships, and multi-salient object scenarios.
Alongside HRSSD, we propose the Deep Spectral Saliency Network (DSSN), establishing a foundational baseline for HRSI-SOD. The three components of DSSN are designed to address aforementioned challenges, collectively enabling accurate and robust target detection.
Experimental results demonstrate the necessity of HRSSD and highlight the effectiveness of DSSN in identifying salient objects across complex scenes.

\section*{Acknowledgments}
This work was financially supported by the National Key Scientific Instrument and Equipment Development Project of China (No. 61527802).

\bibliographystyle{IEEEtran}
\bibliography{ref}

\end{document}